\documentclass[lettersize,journal]{IEEEtran}
\usepackage{amsmath,amsfonts}
\usepackage{algorithmic}
\usepackage{array}
\usepackage[caption=false,font=normalsize,labelfont=sf,textfont=sf]{subfig}
\usepackage{textcomp}
\usepackage{stfloats}
\usepackage{float}
\usepackage[hidelinks]{hyperref}
\usepackage{verbatim}
\usepackage{graphicx}
\usepackage{soul}
\def\BibTeX{{\rm B\kern-.05em{\sc i\kern-.025em b}\kern-.08em
    T\kern-.1667em\lower.7ex\hbox{E}\kern-.125emX}}
\usepackage{lineno}
\usepackage{booktabs}
\usepackage{makecell}
\usepackage{multirow}
\usepackage{color}
\usepackage[table]{xcolor}
\usepackage{balance}
\usepackage{newtxtext}  
\usepackage{newtxmath}  

\definecolor{myyellow}{rgb}{1.0, 0.898, 0.769}

\begin{document}
\title{DiffCom: Decoupled Sparse Priors Guided Diffusion Compression\\ for Point Clouds}

\author{Xiaoge Zhang, Zijie Wu, Mingtao Feng, Mehwish Nasim, Saeed Anwar, Ajmal Mian
\thanks{Xiaoge Zhang, Zijie Wu, Mehwish Nasim, Saeed Anwar, and Ajmal Mian are with the University of Western Australia, Crawley, Perth, WA 6009, Australia (Email: xiaoge.zhang@research.uwa.edu.au; wuzijieeee@hnu.edu.cn; mehwish.nasim@uwa.edu.au; saeed.anwar@uwa.edu.au; ajmal.mian@uwa.edu.au); Mingtao Feng is with Xidian University, Xi'an 710071, China (Email: mintfeng@hnu.edu.cn).}}




\maketitle

\begin{abstract}
Lossy compression relies on an autoencoder to transform a point cloud into latent points for storage, leaving the inherent redundancy of latent representations unexplored. To reduce redundancy in latent points, we propose a diffusion-based framework guided by sparse priors
that achieves high reconstruction quality, especially at low bitrates. Our approach features an efficient dual-density data flow that relaxes size constraints on \emph{latent points}. It hybridizes a Probabilistic conditional diffusion model to encapsulate essential details for reconstruction within \emph{sparse priors}, which are decoupled hierarchically into intra and inter-point priors. Specifically, our DiffCom encodes the original point cloud into latent points and decoupled sparse priors through separate encoders. To dynamically attend to geometric and semantic cues from the priors at each encoding and decoding layer, we employ an attention-guided latent denoiser conditioned on the decoupled priors. Additionally, we integrate the local distribution into the arithmetic encoder and decoder to enhance local context modeling of the sparse points. The original point cloud is reconstructed through a point decoder. Compared to state-of-the-art methods, our approach achieves a superior rate-distortion trade-off, as evidenced by extensive evaluations on the ShapeNet dataset and standard test datasets from the MPEG PCC Group. 
\end{abstract}


\section{Introduction}

\IEEEPARstart{P}{oint} clouds are widely used for representing objects or scenes in various applications such as 3D reconstruction~\cite{yin2025sma, guo2024motion, wu2023sketch}, augmented/mixed reality~\cite{skirnewskaja2024accelerated, sandstrom2023point-mixR}, autonomous driving~\cite{yang2024visual-autodrive, lu2024lidar},  metaverse~\cite{yang2024digitalization}, and industrial production~\cite{liu2025point}. 
Point clouds efficiently represent 3D geometry by storing only surface samples, avoiding voxel-like memory waste in sparse scenes~\cite{liu2019point,qiPointNetDeepLearning2017}. Moreover, the unstructured nature of point clouds, characterized by the absence of fixed connectivity, avoids topological constraints and enables flexible and direct geometric editing~\cite{wang2020cascaded}. With the rapid advancement of 3D acquisition technology, complex geometries can now be captured in real time. However, this poses considerable challenges for storing and transmitting such large volumes of data. For instance, each point cloud in the Owlii dataset ~\cite{owlii} typically contains over 2.5 million points and occupies approximately 150MB of storage, highlighting the need for more efficient compression techniques that significantly reduce storage footprint without compromising geometric fidelity.

\begin{figure}[t]
    \centering
    \includegraphics[width=\linewidth]{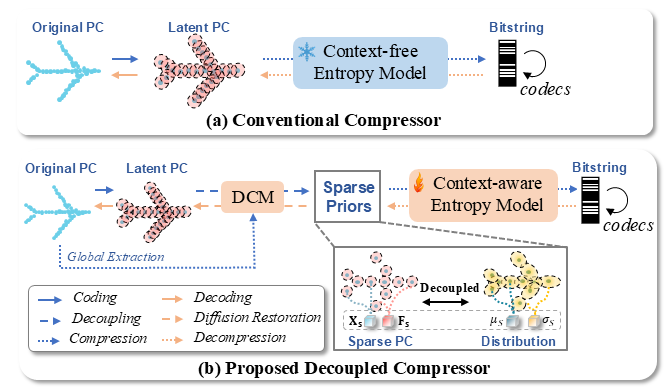}
    \caption{
    (a) Conventional compressors code the latent representation of points directly via a naive context-free entropy model. (b) Our method codes the sparse priors via a context-aware entropy model. It employs a two-stage data flow to compress the 
    points further into decoupled sparse priors. It incorporates a Probabilistic attention-based conditional diffusion model to denoise the latent representations conditioned on the sparse priors.}
\label{fig:quickrun}
\end{figure}

Lossy geometry employs an encoder to map a point cloud into low-dimensional latent points as well as 
associated features, which we refer to as \textit{latents} briefly unless otherwise specified. The latents are then coded into a binary string using an entropy model for storage, and the decoder reconstructs the original point cloud from the decoded latents. 
Voxel-based methods~\cite{wang2021lossy, nguyenLearningBasedLosslessCompression2021} transform point clouds into voxels before applying volumetric encoders to acquire latent codes. While these methods can be easily extended from well-established image compression techniques, they suffer from significant computational overhead and geometric loss due to voxelization. Despite attempts to reduce memory~\cite{wang2021multiscale} by using sparse convolution~\cite{choy20194d}, these networks struggle with high sparsity, resulting in inadequate local context. Point-based methods typically utilize an auto-encoder structure, as shown in Fig.~\ref{fig:quickrun}(a). 
\cite{yan2019deep-autoencode1} and~\cite{huang20193d-autoencode2} propose to directly compress the raw point cloud data in the latent space, with the latent representations extracted by existing backbone networks such as  PointNet~\cite{qiPointNetDeepLearning2017} or PointNet++ 
\cite{qiPointNetDeepHierarchical2017}). However, the decompressed point clouds tend to lose local details, making it difficult to balance the compression rate and reconstruction quality. Recently, he~\emph{et~al.}~\cite{he2022density} proposed a density-aware encoder to retain local density by enforcing density loss. However, as the compression ratio increases, this method suffers from rapid degradation of local contextual information, imposing a lower bound on the latent redundancy required to maintain acceptable quality.
\begin{figure}[tbp]
    \centering
    \includegraphics[trim=0cm 8cm 12cm 0cm, width=\linewidth]{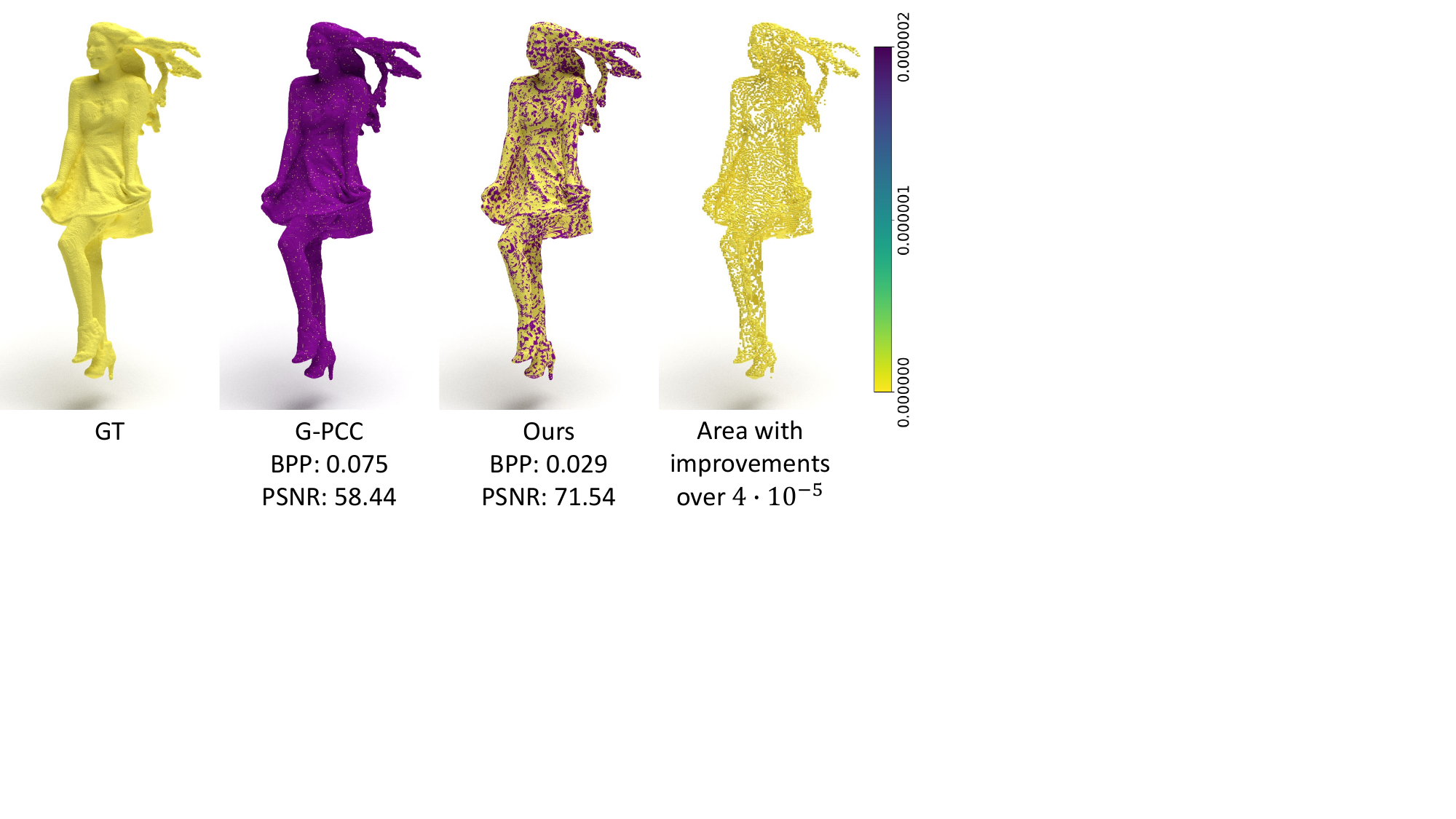}
    \caption{Improvements achieved by the proposed method. Errors beyond the maximum threshold are truncated. All errors are computed from the ground truth to the reconstructed point cloud and visualized on the ground truth point cloud.}
\label{fig:improve}
\end{figure}
To address redundancy in latent representations, we propose a dual-density data flow, as illustrated in Fig.~\ref{fig:quickrun}(b). The data  flow separates the functions of reconstruction and storage into two complementary components: dense latents and sparse priors. These latent representations, generated by PointNet-like encoders, consist of point subsets with associated features. The latent points capture an essential global structure, while their features abstract the local context of surrounding points. Previous work has shown that such representations often include redundant points sampling the same surface regions. To address this issue, we extract a sparser subset of latent points, referred to as sparse points, using a higher sampling rate. However, naive dimensionality reduction inevitably results in contextual information loss and fails to capture spatial dependencies among sparse points. To overcome this, we model each sparse point as a minimal anchor unit, representing a localized region characterized by surface geometry, curvature, and point density. A local distribution predictor is introduced to learn adaptive contextual information for each anchor region in a statistically explicit manner. We refer to the combination of the sparser latent subset and its associated local distributions as sparse priors, which are physically encoded and stored. These sparse priors enable more efficient compression and support context-aware binary coding.

A key challenge in faithful reconstruction from sparse priors lies in recovering high-resolution latent representations from limited observations. This problem is twofold: (1) generating dense, high-fidelity latent features from minimal sparse inputs, and (2) effectively leveraging local statistical distributions to guide the reconstruction process. To address the first challenge, we adopt diffusion models due to their strong generative capabilities. Diffusion models have demonstrated strong generative capability across both 2D~\cite{wang2025denoising, wang2025uniadapter, que2024denoising, dhariwal2021diffusion} and 3D domains~\cite{chen2025discrete, shan2024diffusion, luo2021diffusion}, effectively mitigating common issues such as mode collapse and over-smoothing. Crucially, diffusion enables the synthesis of dense latent representations from pure noise, making it more intuitive to inject sparse priors as conditioning signals rather than attempting to directly upsample sparse inputs to a high-dimensional latent space. For the second challenge, we introduce an attention-based aggregation mechanism that fuses sparse observations with local distribution cues. The resulting distribution parameters are further incorporated into a context-aware entropy model, allowing for efficient compression while maintaining high semantic fidelity in the reconstruction.

By employing two distinct encoding streams and utilizing the diffusion compression model as an intermediary, our framework separates the latents used for reconstruction from the sparser representations used for storage. This separation enables more flexible size constraints on the latents while ensuring high reconstruction fidelity, even under high compression ratios. Additionally, to better exploit the intra-point priors, we integrate the local distribution into a context-aware entropy model, thereby improving the binary encoding of sparse points. This approach contrasts with the conventional context-free entropy model, which assumes a global distribution.

Our main contributions can be summarized as follows:
\begin{itemize}
    \item We propose a lossy point cloud compression data flow that compresses a point cloud further into \textit{sparse priors}. By capitalizing on the form simplicity of sparse priors, we relax the size constraints imposed on latents for reconstruction and reduce the inherent redundancy. 
    
    \item We introduce a multi-stage sparsity-guided conditional diffusion scheme. We decouple the sparse priors into inter-point sparse latents and intra-point local distributions, and employ cross-attention to aggregate priors from both scales across varying down-sampling rates. The sparse mapping is learned jointly with the denoising process to preserve consistent topological and semantic properties with the original point cloud.
    
    \item We present a decoupled sparse priors guided diffusion compression model for point cloud, driven by a conditional diffusion model and a context-aware entropy model. Our framework denoises the latent representations conditioned on the sparse priors, enabling the generation of high-quality distribution-enhanced structures even at extremely low bitrates. Our inter-point local distribution is also integrated into both the arithmetic encoder and decoder to enhance the contextual modeling of sparse points.
    
\end{itemize}

\section{Related Work}
This section highlights recent advances in point cloud compression techniques and the application of diffusion models for processing point cloud data.

\noindent\textbf{Point Cloud Compression}. mainly falls into three categories: Voxel-based, Octree-based, and Point-based. 

\vspace{1mm}\noindent\textit{Voxel-based methods.} VoxelDNN~\cite{nguyenLearningBasedLosslessCompression2021} and MSVoxelDNN~\cite{nguyenMultiscaleDeepContext2021} 
first voxelizes and partitions the input point cloud, then feeds the cubes into a 2D convolutional neural network. To reconstruct the original point cloud, the topological relationship between the cubes is also stored using negligible bits.
VoxelDNN~\cite{nguyenLearningBasedLosslessCompression2021} uses a ResNet backbone to predict the occupancy probability distribution of voxels, which suffers from slow encoding and decoding speeds because both processes work sequentially. To address this issue, MSVoxelDNN~\cite{nguyenMultiscaleDeepContext2021} proposes to predict the distribution in a coarse-to-fine order. Lossy compression methods~\cite{quachLearningConvolutionalTransforms2019, wangLearnedPointCloud2021} treat the mapping from latent representations to original voxels as a binary classification problem. Quach et al.~\cite{quachLearningConvolutionalTransforms2019} develops a 3D auto-encoder with an analysis transform, an entropy model for binary compression, and a synthesis transform, which now serves as a standard pipeline for lossy voxel-based compression. Wang et al.~\cite{wang2021lossy} proposed predicting the mean and variance of latent representations to improve coding contexts. Voxel-based methods struggle with unequal voxel occupancy in sparse point clouds and have high overhead with dense point clouds. In contrast, our method directly compresses point sets, increasing scalability for dense point clouds while lowering sensitivity to sparsity.

\vspace{1mm}\noindent\textit{Octree-based methods.} transform a point cloud into an octree with the predetermined number of layers, then use an MLP-based network to predict the distribution of non-leaf nodes. Since each non-leaf node has eight children, the compression can be viewed as a classification into 256 categories. OctSqueeze~\cite{huangOctSqueezeOctreeStructuredEntropy2020} was the first to introduce learnable octree-based point cloud compression. It employs multiple MLP layers to predict the occupancy probabilities of child nodes based on information from parent or ancestor nodes. By encoding nodes layer by layer, the method allows for the simultaneous encoding of multiple nodes within each layer.
Similar in concept to OctSqueeze, OctAttention~\cite{fuOctAttentionOctreeBasedLargeScale2022}  improves node context by incorporating information from nodes within the same layer, achieving high accuracy in probability distribution predictions. 
However, it experiences slow decompression speeds because non-leaf nodes depend on their siblings despite the use of local window parallelism. EHEM~\cite{song2023efficient} is proposed as a solution to this problem. EHEM splits the nodes into two disjoint groups; the first is independent of sibling nodes, and the second group relies on the first group for information from the same layer. This enables the simultaneous inference of all nodes within a group strategy, significantly reducing the decoding time. Compared to voxel-based methods, octree-based methods are more efficient in terms of runtime and memory usage. However, they remain sensitive to point distribution and are less suited for dynamic contexts due to the high maintenance costs associated with hierarchical structures.

\vspace{1mm}\noindent\textit{Point-based methods.} D-PCC~\cite{he2022density} directly process the geometry by an auto-encoder~\cite{hinton2006reducing}. The encoder reduces the point cloud resolution via down-sampling and encodes it into a compact binary format using a bottleneck entropy model. The decoder then reconstructs the full point cloud from the compressed format. The method's performance degrades at high compression ratios due to substantial loss of neighboring information. To address this, we introduce a decoupled data flow that preserves sparse points while employing latent points for point cloud reconstruction. This approach effectively generates high-quality, distribution-enhanced structures, even at extremely low bitrates.

\vspace{1mm}\noindent\textbf{Diffusion for Point Cloud Analysis}. The diffusion model was initially proposed for image generation~\cite{ho2020denoising, song2020denoising} and has progressively been applied to various point cloud analysis tasks in recent years. Luo et al.~\cite{luo2021diffusion} is the first to leverage a diffusion model for unconditional point cloud generation.~\cite{zhou20213d} designs a neural network that combines point and voxel representations for point cloud completion. Wu et al.~\cite{wu2023sketch} 
proposes utilizing sketches and texts as conditions to generate object-level colored point clouds. This method involves fusing sketch and point cloud features through a cross-attention mechanism, with separate diffusion models generating geometry and appearance information. Point-E~\cite{nichol2022point} employs diffusion models to create three-dimensional shapes from images by fine-tuning GLIDE models with specific language prompts. Lyu et al.~\cite{lyu2023controllable} resorts to the diffusion model to generate object-level meshes. The first diffusion model generates a dense point cloud conditioned on a sparse point cloud, whereas the second generates the corresponding features from a dense point cloud. After feeding a point-wise auto-encoder with both geometry and attribute data to reconstruct the point cloud, surface reconstruction is used to generate the mesh. In our approach, diffusion models capture complex data distributions and reduce noise, effectively mitigating data loss associated with high compression rates.

\section{Our Proposed Approach}

\begin{figure*}
\centering
\includegraphics[width=\linewidth, trim=0cm 12.5cm 9cm 1cm]{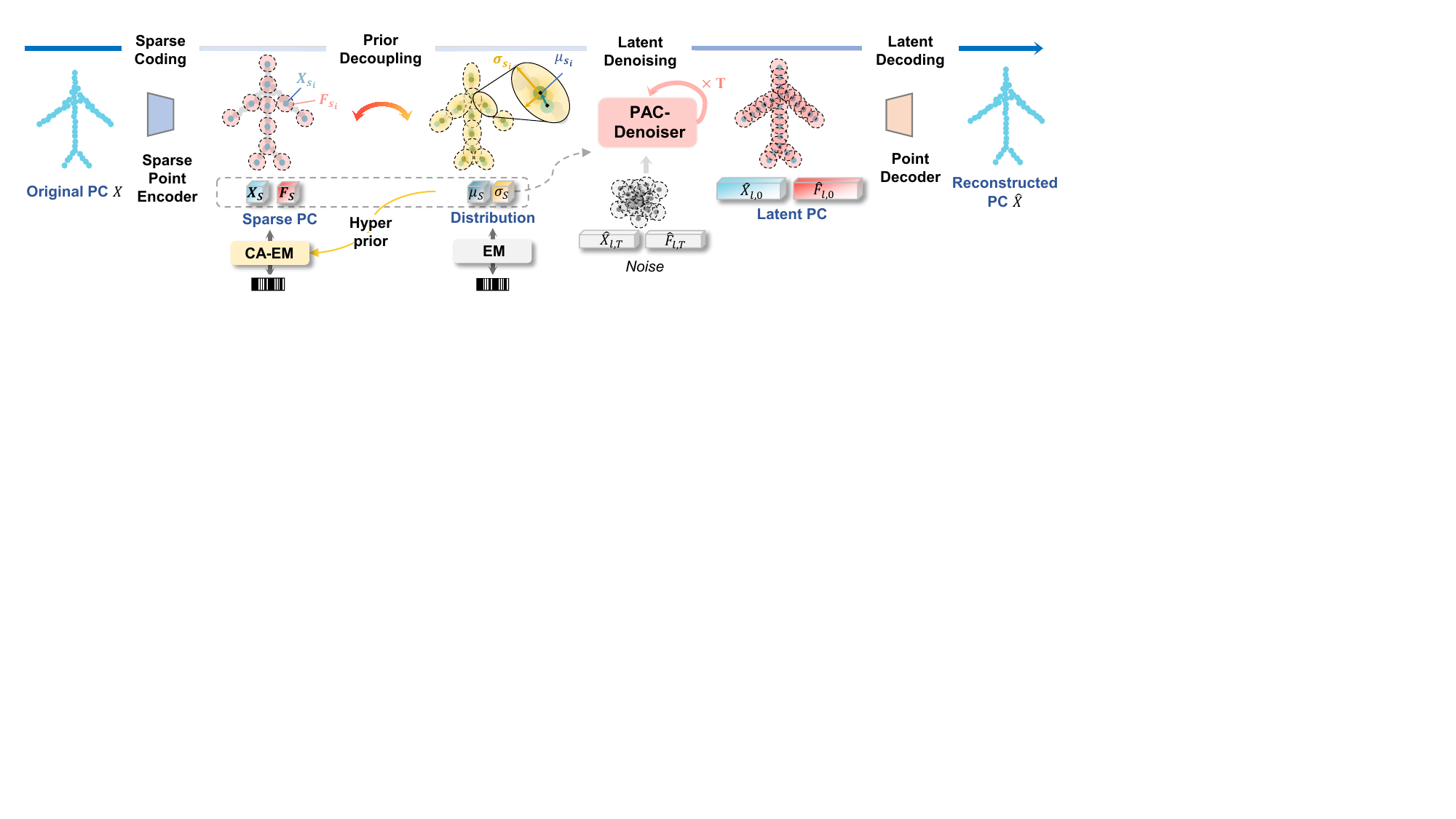}
\caption{Overview of our decoupled sparse priors guided diffusion compression model for point cloud (DiffCom). Instead of directly encoding the input point cloud into latent points and features, we utilize a sparse point encoder to extract a sparser representations. The sparser representation are decoupled into sparse points and intra-point local distributions. During decompression, we begin with Gaussian noise and apply a Probabilistic attention-based conditional denoiser (PACD) conditioned on the sparse priors, reconstructing the latent representation. These reconstructed latents are then decoded to produce a high-quality point cloud.}
\vspace{-5pt}
\label{fig:ppl}
\end{figure*}

A common strategy for encoding a point cloud~$X$ is to use a latent point encoder $\Theta_L$, built from a series of Set Abstraction (SA) modules~\cite{he2022density, lyu2021conditional}. These modules progressively downsample the point set using Farthest Point Sampling (FPS) and aggregate neighboring features to a latent representation. This process can be formally expressed as:
\begin{equation}
    (X_l, F_l) = \Theta_L(X)
\end{equation}
where $X_l=\{x_{l,i}\in{\mathbb{R}^3}|1\leq{i}\leq{N_l}\}$ represents the coordinates of the downsampled points and $F_l=\{f_{l,i}\in{\mathbb{R}^3}|1\leq{i}\leq{N_l}\}$ their corresponding learned features, which describes the local pattern centered at the downsampled points. We call the latent representation $(X_l, F_l)$
as latent point cloud.

However, this process introduces significant \textit{geometric redundancy} as FPS is agnostic to the underlying geometry, for instance, by sampling multiple points from a single continuous surface. This leads to an inefficient representation with highly correlated features, adding unnecessary computational overhead. To address this, we propose a dual-data flow, which can recover the latent point cloud with minimal loss based on a more compact representation (sparse priors). We achieve high compression ratio by storing sparse priors instead of latent points. 
To ensure reconstruction quality, we propose a conditional denoiser that robustly reconstructs the latent point cloud, conditioned on a limited set of sparse priors. Additionally, we propose a context-aware entropy model inspired by derived sparse priors to help binary coding. As depicted in Fig.~\ref{fig:ppl}, our proposed DiffCom separately encodes latent representations and sparse priors, augmented by a latent diffusion model guided by the sparse priors. 

In the subsequent sections, we provide detailed information on the overall data flow, the conditional denoiser, the entropy model, and the model architecture of our proposed framework.
\subsection{Decoupled Sparse Prior Guided Data Flow} \label{subsec:flow}
\vspace{1mm}\noindent\textbf{Sparse coding.} 
 \noindent To effectively compress the original point cloud $X$, we address the inherent redundancy within latent points $X_l$ and associated features $F_l$ by introducing a dual-density data flow. Specifically, we aim to use a sparse point encoder to extract a high-level abstraction from the original point cloud $X$. Our sparse point encoder summarizes an input point cloud into sparse points $X_s=\{x_{s,i}\in{\mathbb{R}^3}|1\leq{i}\leq{N_s}\}$ , which captures essential skeleton, and associated features $F_s=\{f_{s,i}\in{\mathbb{R}^3}|1\leq{i}\leq{N_s}\}$, which capture semantic properties like curvature and regional density within a large receptive field:
\begin{equation}
    (X_s,F_s)=\Theta_S(X)
\end{equation}
where $N_s\ll{N_l}<{N}$. 
 The sparse point encoder $\Theta_S$ resembles $\Theta_L$ architecturally, but employs a higher sampling rate. Their training objectives and ultimate goals diverge significantly: The point encoder $\Theta_L$ is trained with point decoder to directly recover the original point cloud, aiming for a stable latent representation capable of \textit{high-fidelity reconstruction}. In contrast, the sparse point encoder is trained jointly with the denoiser, not for direct reconstruction, but to capture the essential skeletons needed to recover a latent representation. The sparse point cloud yields a coarse approximation of the overall shape, akin to fitting a shape template.

 For simplicity, we will use \(\mathbf{X}\) to denote the pair $(X, F)$ for both sparse and latent point cloud in the remainder of the discussion unless otherwise specified. By storing the sparse coded point cloud instead of the latent point cloud, the compression ratio is significantly improved.
 

\vspace{1mm}\noindent\textbf{Sparse Prior decoupling.} 
The output of our sparse point encoder, the subset $(X_s,F_s)$
, suffers from two main issues: the point locations only coarsely represent the original geometry, and its associated features
are ambiguous "black-box" mixtures of semantics. To overcome these limitations, we treat the sparse points as anchors to build a more accurate and meaningful representation. Our goal is to refine the new representation for high-fidelity latent denoising. We choose the Gaussian Mixture Model (GMM) for our task inspired by its proven success in point cloud registration. In registration, a GMM excels at representing a discrete set of points as a continuous, probabilistic surface. By modeling our sparse points as a GMM, we create a lightweight surrogate that explicitly bridges the information gap without incurring significant storage overhead.

In our framework, the point cloud is represented as a probabilistic mixture of anchors:
\begin{equation}
    p(x)=\Sigma_{i=1}^{N_s}w_i(X_s,Y_s)p_i(x\vert \mu_{s_i},\sigma_{s_i})
\end{equation}
Here GMM is initialized by setting the mean of each Gaussian component to the corresponding sparse point $\mu_{s}^0=X_s$, and  the means are then iteratively refined by $\mu_{s_i}=X_{s_i}+v_i$, the covariance matrix $\sigma_{s_i}$ captures anisotropic properties of local surface, such as the principal directions and degree of stretching. 

Expectation-Maximization (EM) algorithm is used to enforce the distribution converge to the real distribution of the input point cloud. Instead of iterative updates, our model learns to execute the clustering and parameter updating in a single forward pass of a lightweight predictor:
\begin{equation}
    (\mu_s,\sigma_s) = \Theta_P(X_s,F_s)
\end{equation}
The plug-in distribution predictor $\Theta_P$ is implemented as a stack of convolutional layers, yielding a computationally lean and efficient architecture.

The likelihood of the entire original point cloud $L=\Pi_{i=1}^Np(x_i|\mu_s,\sigma_s$) is optimized in the bottleneck, detailed in the loss function section. The calculation of $w_i$ is also learned during the decoding of latent point cloud and will be detailed in the context model section. The GMM-based representation has two advantages. First, their probabilistic nature directly addresses the ambiguity of the features by modeling the underlying data distribution. Second, by representing the data as a continuous surface, GMM is invariant to point cloud resolutions, allowing us to flexible choose sampling rate for point encoder.

Each point, together with its associated features and distribution, forms a highly informative intelligent primitive, which we term \textit{sparse priors}. These priors integrate both handcrafted and learned cues, providing a rich and decoupled representation of the underlying structure. They are then used as conditioning signals for our PACD model, described in the next section. The resulting predicted latents $\tilde{\mathbf{X}}_l$ are subsequently passed to a point decoder to reconstruct the final high-quality point cloud $X$. 

 \subsection{Probabilistic Attention-based Conditional Denoiser}
 \label{subsec:pacd}

 \begin{figure}[tbp]
\centering
\includegraphics[trim=0cm 7.5cm 18cm 0cm, clip, width=\linewidth]{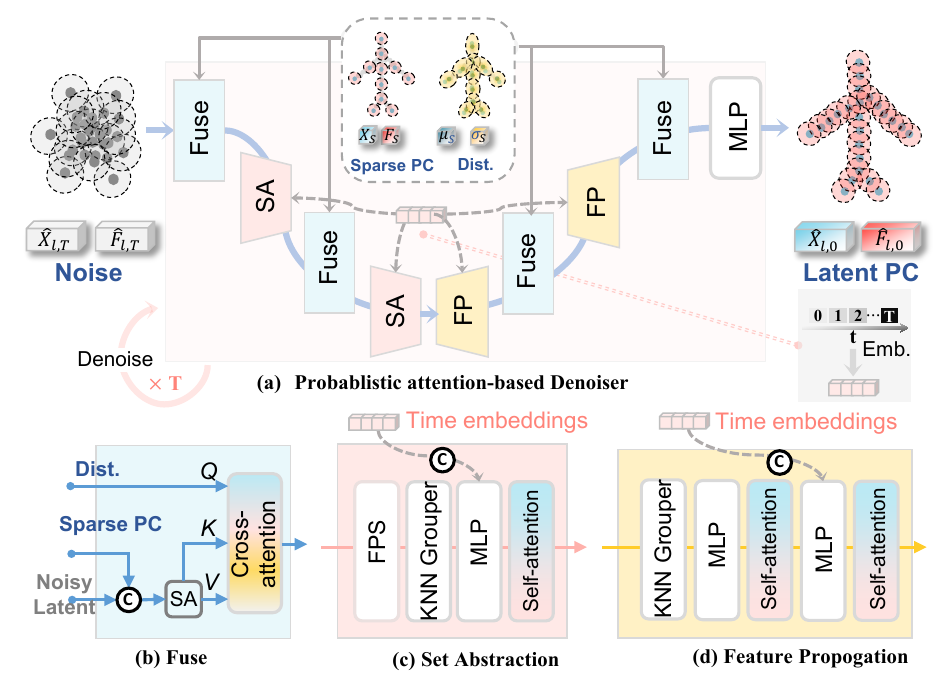}
\caption{Network architecture of PCA-Denoiser. The model adopts a U-Net–like design, where the encoder consists of fusion and set abstraction (SA) modules, and the decoder comprises feature propagation (FP) and fusion modules. An MLP layer maps the decoded features to the target dimension. Time embeddings are injected into both the encoder and decoder to provide timestep-dependent guidance throughout the denoising process.}
\label{fig:denoiser}
\end{figure}
Given the inevitable information loss due to the aggressive down-sampling rate, the mapping from sparse points to latent points becomes ill-posed and prone to clustering artifacts in the reconstructed point cloud. To address this, we utilize diffusion models~\cite{ho2020denoising}, renowned for their robust generative capabilities, to encapsulate the inherent variability in the original point cloud $X$, thereby avoiding mode collapse while reconstruction~\cite{luo2021diffusion}.
The denoiser helps not only enables high-quality reconstruction, but also mitigating inherent outliers through iterative denoising steps. 

To maintain high fidelity of the latents, our DCM incorporates sparse priors into the PACD at each encoding and decoding stage throughout the denoising process. The reverse process of diffusion is modeled as a Markov chain, given by: 
\begin{equation}
p_{\theta}(\mathbf{X}_{l,0},\cdots,\mathbf{X}_{l,T-1}|\mathbf{X}_{l,T})=\prod_{t=1}^{T}p_{\theta}(\mathbf{X}_{l,t-1}|\mathbf{X}_{l,t},\mathbf{X}_s).
 \end{equation}
The stepwise denoising follows a Gaussian distribution:
    $p_{\theta}(\mathbf{X}_{l,t-1}|\mathbf{X}_{l,t},\mathbf{X}_s)$
and the mean $\mu_{\theta}(\mathbf{X}_{l,t},\mathbf{X}_s,t)$ is reparameterized as: 
\begin{equation}
\mu_{\theta}(\mathbf{X}_{l,t},\mathbf{X}_s,t)=\frac{1}{\sqrt{\alpha_t}}(\mathbf{X}_{l,t}-\frac{\beta_t}{\sqrt{1-\bar{\alpha}_t}}\epsilon_\theta(\mathbf{X}_{l,t},\mathbf{X}_s,t)).
\end{equation}
 The noise is encapsulated by the denoising network $\epsilon_\theta$, and $\alpha_t$ and $\beta_t$ are predefined small positive constants, and the cumulative product $\bar\alpha_{t}$.
 
As illustrated in Fig.~\ref{fig:denoiser}, our PACD employs a U-Net-like architecture. At k-th layer, the input is an intentionally noised latent point cloud $(X_l^k, F_l^k)$ based on a pre-scheduled timestep. Prior to each encoding and decoding layer is a fusion layer aiming to fuse the sparse point cloud as guidance. Here we introduce fusion layer specifically. 

Given the sparse point cloud fully as $(X_s, F_s)$. For point locations $X_s$, we safely concatenate them along the point dimension given that sparse points $\mathbf{X}_s^k$ and latents $\tilde{\mathbf{X}}_L^k$ are both proper subsets of the original point cloud and share homogeneous characteristics. For associated features, as the number of channels of $F_s$ differ from that of $F_l$, we use several MLPs to align the feature dimensions.  

After obtaining concatenated inputs $(\hat{X}_l^k,X_s^k)$, we use an MLP layer to extract point-wise features. Then we rely on an improved set abstraction layer~\cite{lyu2021conditional} to perform region-wise feature aggregation. This results in a coarse prediction $\tilde{\mathbf{F}}_l^{k-1}$ for $(k-1)_{th}$ layer, formulated as:
\begin{equation}
    \hat{F}_l^{k-1,\mathrm{crs}}=\mathbb{S}(\Phi((\hat{X}_l^k\Vert\hat{X}_s^k))),
\end{equation}
Here, $\mathbb{S}(\cdot)$ denotes the set abstraction function, $\Phi(\cdot)$ represents a Multi-Layer Perceptron (MLP) transformation, and $\Vert$ is the concatenation operator.
 We hypothesize that local distribution among the sparse points captures essential inter-point contextual information, which plays a pivotal role in enhancing the accuracy of our coarse predictions. 
 We employ a cross-attention mechanism to integrate it with the coarse latent for scalability and explicitness. Following a linear transformation, sparse priors are integrated using an attention mechanism, where the coarse prediction serves as the query, the sparse point cloud as the key, and the inter-point distribution as the value:
 \begin{equation}
\tilde{\mathbf{F}}_l^{k-1} = \chi\left( \phi_q(\tilde{\mathbf{F}}_l^{k-1,\mathrm{crs}}),\, \phi_k(\mathbf{X}_s^{k-1}),\, \phi_v(\boldsymbol{\mu}^{k-1}, \boldsymbol{\sigma}^{k-1}) \right)
\end{equation}
where $\chi(Q,K,V) = \mathrm{softmax} \left( \frac{QK^\top}{\sqrt{d}} \right)$. The attention-based fusion layer enables us to propagate the sparse distribution to latents, yielding a distribution-enhanced output $\tilde{\mathbf{F}}_l^{l-1}$. 

Through iterative sparse guided downsampling and attention-based fusion, we obtain refined features for the latents. The decoder part of the denoiser follows a similar scheme, with the set abstraction replaced by the feature mapper layer. The details of the set abstraction and feature mapping layers are provided in a later section.

 \subsection{Context-aware Entropy Model} \label{subsec:entropy}
 
 Recall that we derive the local distribution \((\mu_s, \sigma_s)\) upon the sparse points to provide extra guidance in the latent denoising process. We further utilize it to capture the coherent spatial dependencies within \(X_s\) during binarization by a naive bottleneck model.

 Before predicting the local distribution, we first transform the sparse points $X_s$ to analyzed features $Y_s$. After quantization, the resulting $\hat{Y}_s$ serves as the main data to be binary coded, and the estimated bits are lower-bounded by the entropy of $\hat{Y}_s$. Given that the entropy increases with the KL divergence between the true and predicted distributions, we argue that the learned local distribution provides a more accurate approximation of the underlying data statistics than default distribution. Accordingly, we design a context-aware entropy model guided by these learned local distributions to replace the conventional context-free entropy model.
 
 Our distribution predictor consists of an analysis transform $\mathcal{H}_a$, a quantizer, and a synthesis transform $\mathcal{H}_s$. More expressively, the hyperparameter $Z_s$ is obtained by an analysis transform $\mathcal{H}_a: Y_s\mapsto{Z_s}$.
 After being quantized to $\hat{Z}_s$, the distribution parameters are obtained by a synthesis transform $\mathcal{H}_s: \hat{Z}_s\mapsto{(\mu_s,\sigma_s)}$. The quantizer is made differentiable by adding a uniform noise~\cite{balle2016end} during training. For simplicity, we will discard the lowercase $s$, which means sparse samples when discussing the distribution of $\hat{Z}_s$ and $\hat{Y}_s$. 
 
  Inspired by hyperprior-based context model in image compression~\cite{minnen2018joint, balle2018variational} and point cloud compression~\cite{wang2021lossy}, we integrate the learned distribution parameters as priors into non-parametric, fully factorized density model:
 \begin{equation}
p_{\tilde{z}|\xi}(\tilde{z}|\xi)=\prod_i(p_{\tilde{z}_{i}|\xi^{(i)}}(\xi^{i})\ast{\mathcal{U}}(-\frac{1}{2},\frac{1}{2}))(\tilde{z}_{i}).
 \end{equation}
 So the distribution of each analyzed sparse point can be modeled as independently conditioned on the priors, calculated by convolving the $uniform$ distribution introduced by quantization and $Laplace$ distribution.
This gives us the local probability density function of $\hat{Y}_s$:
\begin{equation}
p_{\hat{y}_{i}|\hat{\mu}_{i},\hat{\sigma}_{i}}(\hat{y}_{i}|\hat{\mu}_{i},\hat{\sigma}_{i})=\int_{\hat{y}_{i}-\frac{1}{2}}^{\hat{y}_{i}+\frac{1}{2}}\mathcal{L}(y|\hat{\mu}_i,\hat{\sigma}_i)dy.
\end{equation}
The coding of analyzed sparse points $\hat{Y}_s$ is achieved by an arithmetic coder according to the derived probability function, and the distribution parameters $\hat{Z}_s$ are coded separately by a conventional arithmetic coder.

\subsection{Model Architecture}

\begin{figure*}[tbp]
\center{
\includegraphics[trim=0cm 9cm 9.5cm 0cm, clip, width=0.96\linewidth]{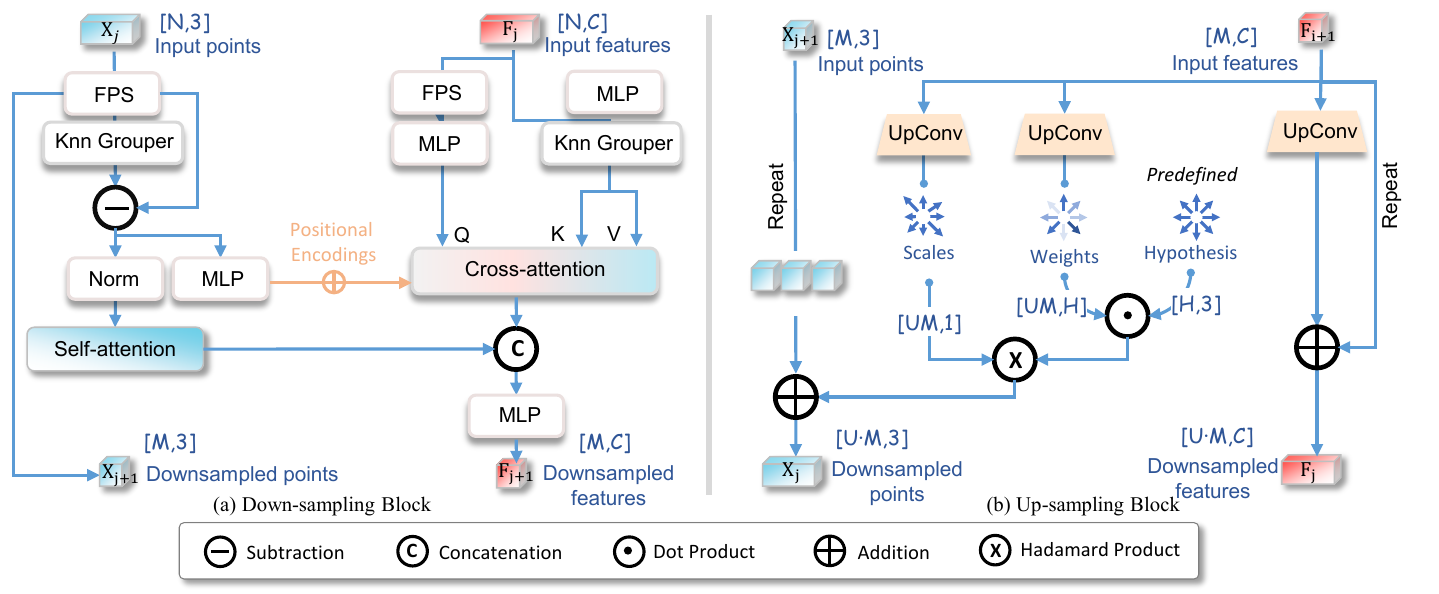}}
\vspace{-4mm}
\caption{Architecture of the downsampling and upsampling blocks. In the downsampling block, points are downsampled using a set abstraction (SA) module, and features are enriched with a PointTransformer layer. In the upsampling block, residual point coordinates are reconstructed by learning scales and weights over predefined directional hypotheses, while residual features are recovered through UpConv operations.}
\label{fig:encoder}
\end{figure*}

\noindent \textbf{Point Encoder.} Our approach initiates with an autoencoder $\Theta_L$ that maps the original point cloud $X=\{x_i\in{\mathbb{R}^3}|1\leq{i}\leq{N}\}$ to latent points $X_l=\{x_{l,i}\in{\mathbb{R}^3}|1\leq{i}\leq{N_L}\}$ and corresponding features $F_{l}=\{f_{l,i}\in{\mathbb{R}^{d}}|1\leq{i}\leq{N_L}\}$:
\begin{equation}
    \Theta_L: X\in{\mathbb{R}^3}\mapsto(X_l, F_l).
\end{equation} After training, we discard the encoder and keep only the decoder for decompression.

In the compression scenario, guidance information is unavailable during decompression, and the use of skip connections between the encoder and decoder is also prohibited (since we do not have the full data during decompression). Therefore, we follow the autoencoder backbone in D-PCC~\cite{he2022density} consisting of downsampling and upsampling stages. Each encoding stage begins by downsampling the point cloud with a predefined downsampling rate using the farthest point sampling algorithm to ensure maximum coverage.

 As shown in Fig.~\ref{fig:encoder}(a), each downsample block integrates explicit local encoding and implicit feature extraction. Initially, the point cloud is mapped to a downsampled version $P_{s+1}$, aggregating the neighbouring information into features $F_{s+1}$. Let us denote the retained key points as $p$ and its neighbouring points in the original point cloud as $C(p)$. For each $p_k\in{C(p)}$, compute the direction from $p$ to $p_k$, and their Euclidean distance, which jointly characterize the local surface variations.
where $p\in{P}_{s+1}$. The geometric variations are further refined via self-gating.
Additionally, a transformer aggregates semantic features from stage $s$ by treating $P_{s+1}$ as the query, $P_s$ as the key, and $F_s$ as the value. Finally, an MLP merges both explicit position embedding and implicit ancestor features from stage $s$.

\vspace{1mm}\noindent \textbf{Point Decoder.} Symmetrically, in the decoding stage $s$, the goal is to upsample points from $(P_s, F_s)$. The upsample block is depicted in Fig.~\ref{fig:encoder}(b). 
Specifically, during the upsampling of $P_{s}$ to generate $\tilde{P}_{s+1}$, a pre-defined number of candidates are defined. To mitigate artifacts in $\tilde{F}_{s+1}$, we also apply a sub-point convolution technique as D-PCC~\cite{he2022density}. For each candidate, direction and scale are predicted from $F_s$, and the weighted sum is computed as the residual for $\tilde{C}(p)$. The Sub-Point Convolution will generate $k>f_s$ upsampled point candidates, from each we select the top $f_s$ points and associated features. Unlike D-PCC, our method uses $f_s$ to control the number of expanded points, rather than a learnable $\tilde{C}(p)$. This is motivated by our empirical observation that the learned upsampling rate consistently converges to the fixed downsampling rate in the encoder.
Following the $s$ upsampling stages, we apply an upsampling module with a rate of 1 for final refinement.

\vspace{1mm}\noindent \textbf{Set Abstraction Module.} Set abstraction (SA) modules~\cite{lyu2023controllable} are utilized to carry out downsampling in the denoiser. For simplicity, we denote the concatenated input $(\tilde{X}_l^k, X_s^k)$ as $X_k=\{x_j|1\leq{j}\leq{N_k}\}$, where $x_j$ denotes the 3D coordinates and $N_k$ is the number of points at level $k$, the associated features are given by $F_k=\{f_j|1\leq{j}\leq{N_k}\}$. The concatenation $(X_k, F_k)$ is of size $N_k \times (d_k + 3)$.

The module starts by using iterative Farthest Point Sampling (FPS) to reduce the number of points to $N_{k+1}$: $\{x_j|1\leq{j}\leq{N_{k+1}}\}$. For each sampled point $x_j$, it identifies $K$ nearest neighbors from the input of the last layer $(X_k, F_k)$. The neighbors' coordinates and features are extracted as matrix of shape $N_{k+1} \times K \times (d_k + 3)$, which is processed by a shared multi-layer perceptron (MLP) to produce a new matrix with dimensions $N_{k+1} \times K \times d_{k+1}$. A self-attention layer is then applied to aggregate the features from $K$ neighbors, yielding the output matrix $F_{k+1}$ of size $N_{k+1} \times d_{k+1}$. 

\vspace{1mm}\noindent \textbf{Feature Propagation Module.} Feature propagation (FP) modules~\cite{lyu2023controllable} are utilized to carry out downsampling in the denoiser. In the feature propagation module, each FP layer is the reverse process of the corresponding SA layer: from \(\{x_j|1\leq{j}\leq{N_{k+1}}\}\) to \(\{x_j|1\leq{j}\leq{N_k}\}\). This is achieved by finding the neighbors \(\{y_n|{n}\in{B_y(x_j)}\} \subseteq \{y_n|1\leq{n}\leq{N_{k+1}}\}\) for each \(x_j\). The features at these neighbors are transformed via a shared MLP and aggregated to \(x_j\) through an attention mechanism. Finally, the upsampled features are concatenated with the skip-connected features from the corresponding SA module and processed through a naive PointNet~\cite{qiPointNetDeepLearning2017}. 

\vspace{1mm}\noindent \textbf{Sparse Point Encoder.} To generate homogeneous conditions for our HMA-guided latent diffusion model, we adopt the same architecture as the point encoder, but with a higher down-sample rate to extract a sparser point cloud.

\subsection{Loss Function}
Firstly, we pre-train the encoder $\Theta_L$ and decoder $\Omega_L$ using the Chamfer Distance (CD) loss only.
Then we train the diffusion model to learn the noise $\epsilon$, and the reconstruction loss is given by:
\begin{small}
    \begin{equation}
\ell_{recon}=\mathcal{E}_{(X_l,X_s)\sim{P_{data}}}\|\epsilon-\epsilon_{\theta}(\sqrt{\bar{\alpha_t}})X_l+\sqrt{1-\bar{\alpha_t}}\epsilon, X_l,X_s)\|^2
\end{equation}
\end{small}
For compression loss $\ell_{comp}$, the Bits Per Point (BPP) is approximated by the likelihood of sparse points and distributions:
\begin{equation}
\ell_{comp}=E_{\hat{Y}_s}[-\log_2(p_{\hat{Y}_s}(\hat{Y}_s)] + E_{\hat{Z}_s}[-\log_2(p_{\hat{Z}_s}(\hat{Z}_s)]
\end{equation}
The final loss of denoiser is a weighted sum of two losses.

\begin{figure}[ht]
\includegraphics[trim=0cm 5cm 13cm 0cm, width=\linewidth]{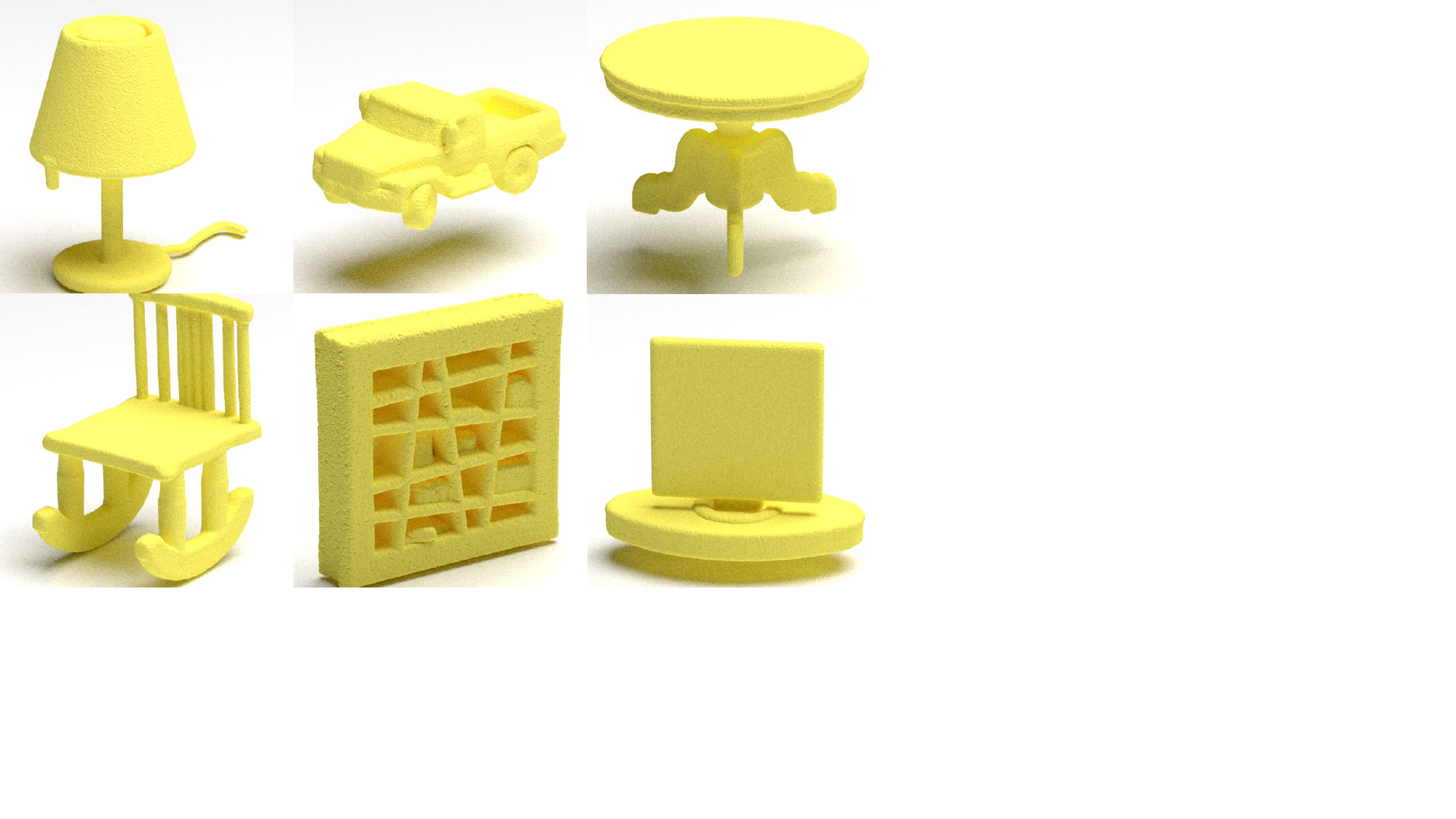}
\caption{Training samples from the ShapeNet dataset.}
\label{fig:sample_sn}
\end{figure} 

\begin{figure}[h]
\includegraphics[trim=0cm 8cm 6cm 0cm, width=\linewidth]{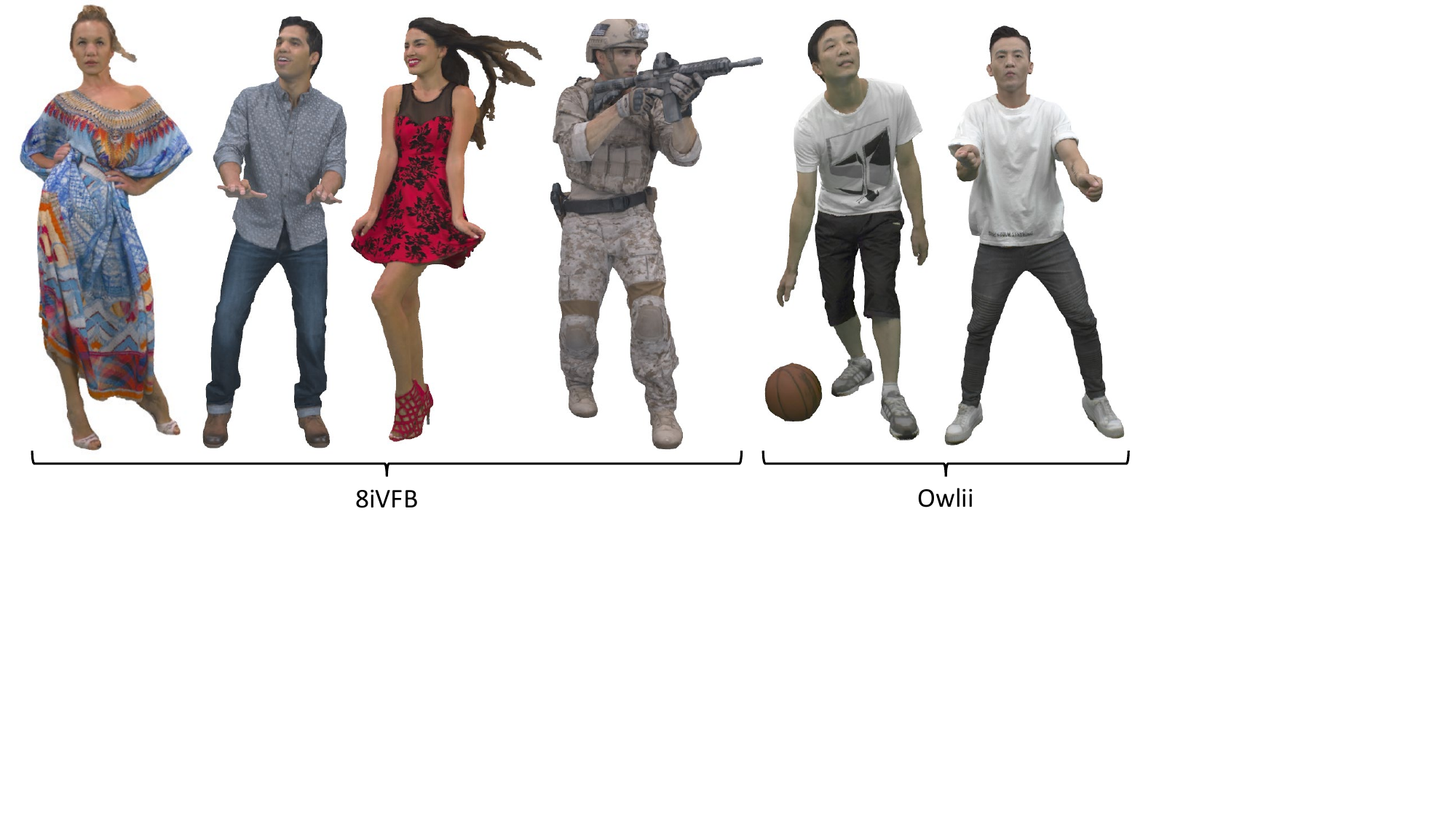}
\caption{Test samples from the 8iVFB and Owlii datasets. From left to right: Longdress, Loot, RedandBlack, Soldier, BasketballPlayer, and Dancer.}
\label{fig:sample_mpeg}
\end{figure} 

\section{Experiments}
\label{sec:experiments}

\subsection{Experimental Setup}
\noindent\textbf{Datasets}. We train our model on the ShapeNet dataset preprocessed by~\cite{peng2021shape}, which contains 30,661 object-level point clouds across 13 categories, using the official training and testing splits. The training samples are visualized in Fig.~\ref{fig:sample_sn}. During training, we sample 2048 points per point cloud, while for testing, we evaluate on $N=2k,10k,32k$ points. Additionally, we evaluate the performance on seven test samples under the test conditions specified by the MPEG Group~\cite{graziosi2020overview}, including \texttt{longdress\_vox10\_1300}, \texttt{loot\_vox10\_1200}, \texttt{redandblack\_vox10\_1200}, \texttt{soldier\_vox10\_0690}, and \texttt{queen\_0200} from the 8iVFB dataset, as well as \texttt{basketball\_player\_vox11\_0200} and \texttt{dancer\_vox11\_0001} from the Owlii dataset. The testing samples are visualized in Fig.~\ref{fig:sample_mpeg}. We collectively refer to the samples as the MPEG dataset.

\vspace{1mm}\noindent\textbf{Evaluation Metrics.}
Following previous methods~\cite{he2022density, wang2022sparse}, we use Chamfer Distance (CD) and PSNR to evaluate geometry accuracy, and Bits Per Point (BPP) to assess the compression ratio. For overall performance across the bitrates, we employ BD-PSNR and BD-Rate~\cite{bjontegaard2001calculation}. 
Note that a lower BPP indicates a higher compression ratio, requiring less space for transmission and storage, while a higher PSNR reflects better reconstruction quality.

\vspace{1mm}\noindent\textbf{Baselines.}
We select five deep-learning-based lossy point cloud compression methods: Depoco~\cite{wiesmann2021deep}, D-PCC~\cite{he2022density}, PCGCv2~\cite{wang2021multiscale}, SparsePCGC~\cite{wang2022sparse}, UniPCGC~\cite{wang2025unipcgc}, and the latest standard test condition TMC13-v23.0~\cite{graziosi2020overview} by the MPEG PCC Group. To ensure a fair comparison, all deep learning methods were retrained on ShapeNet using only geometric coordinates. For evaluation, we use 2k, 10k, and 32k points on the ShapeNet dataset, with 2k points as the default unless otherwise specified. For the 2k-point setting, baseline methods are implemented in sparse mode when supported, while dense mode is adopted for all other experiments.

\begin{figure*}[htbp]
\includegraphics[trim=4.5cm 1cm 4.5cm 1cm, width=\linewidth]{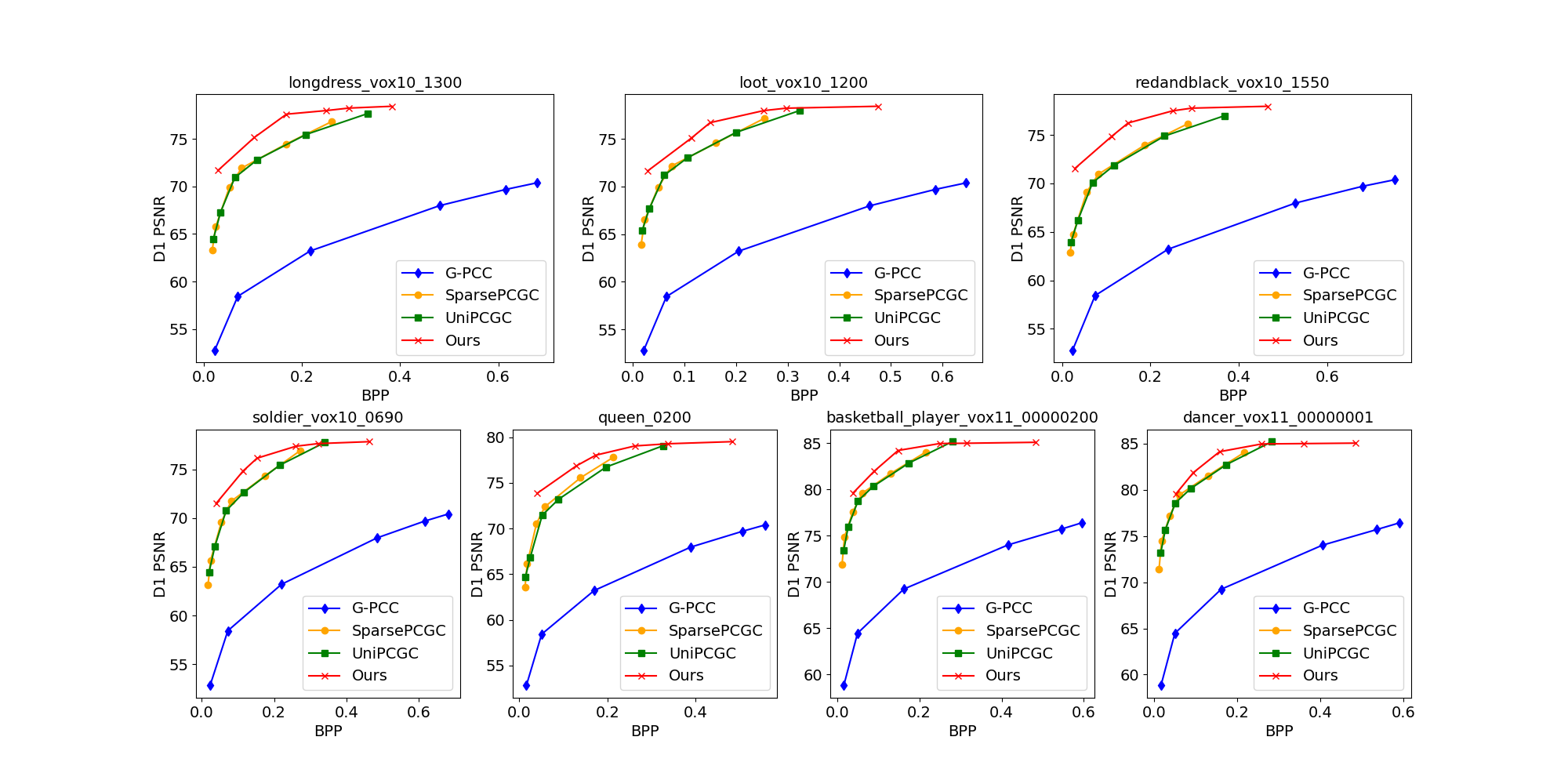}
\vspace{-9mm}
\caption{Rate-Distortion Performance on MPEG dataset. BPP (lower is better) indicates compression rate, and D1 PSNR (higher is better) indicates reconstruction quality.}
\vspace{-3mm}
\label{fig:mpeg_rdcurve}
\end{figure*}

\begin{table*}[htbp]
    \centering
    \caption{Improvements of our method (highlighted in yellow) over three existing methods on MPEG dataset. We also report encoding/decoding time. Our method achieves significant improvements in Rate and PSNR, and has the fastest decoding time.}
    \vspace{-3mm}
    \footnotesize
    \setlength\tabcolsep{3pt}
    \resizebox{0.9\textwidth}{!}{\begin{tabular}{l|cccc|cccc|cccc|c c}
    \toprule
    \multirow{3}{*}{Test Data} &  \multicolumn{4}{c|}{G-PCC~\cite{graziosi2020overview}} & \multicolumn{4}{c|}{SparsePCGC~\cite{wang2022sparse}} & \multicolumn{4}{c|}{UniPCGC~\cite{wang2025unipcgc}} & \multicolumn{2}{c}{Ours} \\
    & \multicolumn{2}{c}{Gain} & \multicolumn{2}{c|}{Time (s)} & \multicolumn{2}{c}{Gain} & \multicolumn{2}{c|}{Time (s)} & \multicolumn{2}{c}{Gain} & \multicolumn{2}{c|}{Time (s)} & \multicolumn{2}{c}{Time (s)} \\

    & Rate ($\downarrow$) & PSNR($\uparrow$) & Enc & Dec
    & Rate ($\downarrow$) & PSNR($\uparrow$) & Enc & Dec & Rate ($\downarrow$) & PSNR($\uparrow$) & Enc & Dec & Enc & Dec \\
    \midrule
    longdress\_vox10\_1300 & \cellcolor{myyellow} -98.52 & \cellcolor{myyellow} +15.06 & 1.48 & 0.61 & \cellcolor{myyellow} -41.18 & \cellcolor{myyellow} +2.69 & \textbf{0.21} & \underline{0.43} & \cellcolor{myyellow} -44.43 & \cellcolor{myyellow} +2.72 & \underline{0.92} & 0.68 & 1.51 & \textbf{0.27}\\
    loot\_vox10\_1200 &  \cellcolor{myyellow} -98.25 & \cellcolor{myyellow} +14.30 & 1.37 & 0.57 & \cellcolor{myyellow} -32.76 & \cellcolor{myyellow} +2.07 & \textbf{0.20} & \underline{0.41} & \cellcolor{myyellow} -33.04 & \cellcolor{myyellow} +2.04 & \underline{0.90} & 0.62 & 1.44 & \textbf{0.26}\\
    redandblack\_vox10\_1200 & \cellcolor{myyellow} -98.36 & \cellcolor{myyellow} +14.72 & 1.30 & 0.54 & \cellcolor{myyellow} -47.91 & \cellcolor{myyellow} +3.25 & \textbf{0.20} & \underline{0.39} & \cellcolor{myyellow} -50.37 & \cellcolor{myyellow} +3.34 & \underline{0.91} & 0.64 & 1.57 & \textbf{0.26}\\ 
    soldier\_vox10\_0690 & \cellcolor{myyellow} -96.94 & \cellcolor{myyellow} +14.03 & 1.86 & 0.77 & \cellcolor{myyellow} -33.68 & \cellcolor{myyellow} +2.13 & \textbf{0.24} & \underline{0.54} & \cellcolor{myyellow} -32.49 & \cellcolor{myyellow} +2.12 & \underline{0.92} & 0.81 & 1.60 & \textbf{0.27}\\
    queen\_0200 & \cellcolor{myyellow} -99.86 & \cellcolor{myyellow} +14.27 & 1.68 & 0.65 & \cellcolor{myyellow} -24.72 & \cellcolor{myyellow} +1.84 & \textbf{0.20} & \underline{0.45} & \cellcolor{myyellow} -31.64 & \cellcolor{myyellow} +2.10 & \underline{0.89} & 0.71 & 1.55 & \textbf{0.28}\\
    basketball\_player\_vox11\_0200 & \cellcolor{myyellow} -99.89 & \cellcolor{myyellow} +14.41 & 5.16 & 1.98 & \cellcolor{myyellow} -32.98 & \cellcolor{myyellow} +1.53 & \textbf{0.48} & \underline{1.32} & \cellcolor{myyellow} -32.62 & \cellcolor{myyellow} +1.49 & \underline{1.23} & 1.85 & 5.42 & \textbf{0.84}\\
    dancer\_vox11\_0001 & \cellcolor{myyellow} -99.48 & \cellcolor{myyellow} +13.93 & 4.56 & 1.75 & \cellcolor{myyellow} -27.58 & \cellcolor{myyellow} +1.33 & \textbf{0.44} & \underline{1.17} & \cellcolor{myyellow} -27.52 & \cellcolor{myyellow} +1.28 & \underline{1.23} & 1.65 & 4.27 & \textbf{0.69}\\
    \midrule
    Average & \cellcolor{myyellow} -98.46 & \cellcolor{myyellow} +14.62 & 2.27 & 0.87 & \cellcolor{myyellow} -34.27 & \cellcolor{myyellow} +2.13 & \textbf{0.27} & \underline{0.74} & \cellcolor{myyellow} -36.52 & \cellcolor{myyellow} +2.12 & \underline{1.01} & 1.00 & 2.13 & \textbf{0.45} \\
    \bottomrule
    \end{tabular}}
    \label{tab: mpeg-rdgain}
    \vspace{-3mm}
\end{table*}

\subsection{Experimental Results}
\noindent\textbf{Rate-Distortion Performance.} Fig.~\ref{fig:mpeg_rdcurve} and Table~\ref{tab: mpeg-rdgain} present the rate–distortion curves and the corresponding performance gains on the MPEG dataset. We compare our method against the latest standard platform G-PCC (v23) and two state-of-the-art voxel-based approaches, UniPCGC~\cite{wang2025unipcgc} and SparsePCGC~\cite{wang2022sparse}. Table~\ref{tab: mpeg-rdgain} summarizes rate savings and PSNR gains per sample and overall. For instance, on \texttt{redandblack\_vox10\_1200}, our method achieves 98.52\% rate savings and a 15.06 dB PSNR gain over G-PCC. On average, our DiffCom achieves 98.46\% rate savings compared with G-PCC, 34.27\% compared with SparsePCGC, and 36.52\% compared with UniPCGC, with corresponding PSNR gains of 14.62 dB, 2.13 dB, and 2.12 dB, respectively. Fig.~\ref{fig:mpeg_rdcurve} shows our method consistently delivers superior reconstruction quality across all bitrates, demonstrating robust improvements across compression levels.
The PSNR improvements at BPP~$<$~0.4 are particularly significant, indicating that our decoupled sparse priors enable the preservation of essential geometric details under aggressive compression. 
For \texttt{basketball\_player\_vox11\_0200} and \texttt{dancer\_vox11\_0001}, our method is slightly outperformed by UniPCGC at BPP~$>$~0.4, as the increasing bits help UniPCGC capture finer details, while our dual data flow has already preserved sufficient structure at lower rates.

\begin{figure}[htbp] \vspace{-3mm}
\includegraphics[trim=0cm 1cm 0cm 0cm, width=\linewidth]{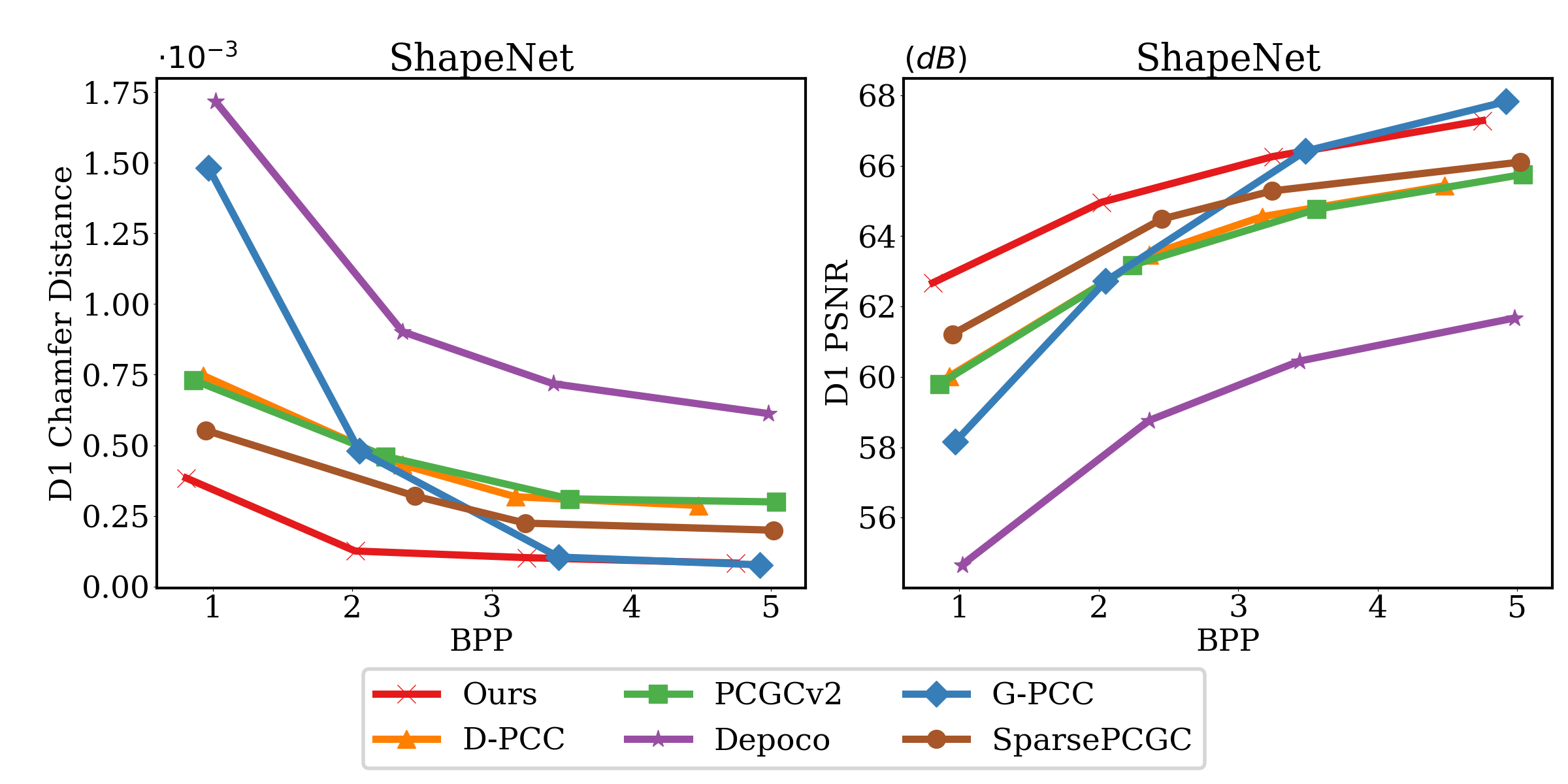}
\vspace{-6mm}
\caption{Rate-Distortion Performance on ShapeNet dataset.}
\vspace{-2mm}
\label{fig:shapenet_rdcurve}
\end{figure} 

On the ShapeNet dataset, we further compare our method against G-PCC (v23)~\cite{graziosi2020overview}, two voxel-based methods, PCGCv2~\cite{wang2021multiscale} and SparsePCGC~\cite{wang2022sparse}, and two point-based methods, Depoco~\cite{wiesmann2021deep} and D-PCC~\cite{he2022density}. The corresponding rate–distortion curves and performance gains are shown in Fig.~\ref{fig:shapenet_rdcurve} and Table~\ref{tab: psnr_sn}. Fig.~\ref{fig:shapenet_rdcurve} demonstrates that DiffCom achieves state-of-the-art performance across almost all compression rates. Table~\ref{tab: psnr_sn} further shows that DiffCom surpasses Depoco with an 89.31\% rate saving and a 6.89 dB PSNR gain.  

\begin{table}[htbp]
\centering
\caption{Bitrate and PSNR gains on the ShapeNet dataset, using Depoco as the anchor. Our method is highlighted. Best results are \textbf{bolded}, and the second-best results are \underline{underlined}.}
\setlength\tabcolsep{14pt}
\footnotesize
\begin{tabular*}{\columnwidth}{l|cc}
\toprule
Methods & PSNR Gains($\uparrow$) & Rate Gains($\downarrow$)  \\
\midrule
Depoco~\cite{wiesmann2021deep} & - & -  \\
PCGCv2~\cite{wang2021multiscale} & +4.72 dB & -69.30\%  \\
D-PCC~\cite{he2022density} & +4.80 dB & -67.47\% \\
SparsePCGC~\cite{wang2022sparse} & \underline{+5.64 dB} & \underline{-76.61\%} \\
Ours & \cellcolor{myyellow} \textbf{+6.89 dB} & \cellcolor{myyellow} \textbf{-89.31\%}  \\
\bottomrule
\end{tabular*}
\label{tab: psnr_sn}
\end{table}

Overall, the results confirm that for both sparse and dense point clouds, our DiffCom effectively maintains key geometric information at low bitrates while ensuring comparable reconstruction quality at high bitrates.


\begin{figure*}[tbp]
\centering
\includegraphics[width=\linewidth, trim=0cm 0cm 0cm 0cm]{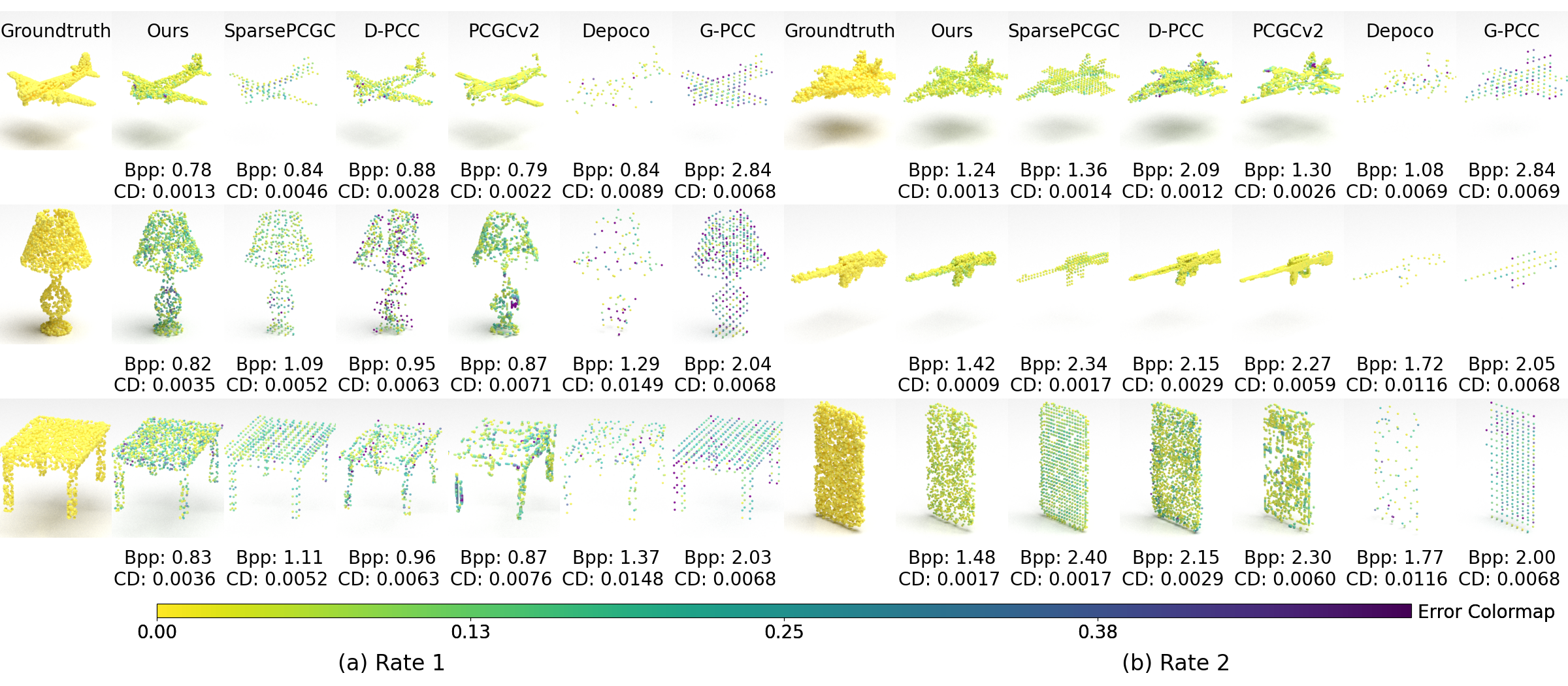}
\caption{Qualitative results on ShapeNet at two selected rates. For fairness, baselines are compared at matched bitrates ((a) BPP $\approx$ 0.9 and (b) BPP $\approx$ 2.0). Errors beyond the maximum threshold are truncated. All errors are computed from the reconstructed to the ground truth point cloud and visualized on the reconstructed point cloud.}
\label{fig:shapenet_vis}
\end{figure*}

\begin{figure}[htbp]
\centering
\includegraphics[width=\linewidth, trim=0cm 8cm 12cm 0cm]{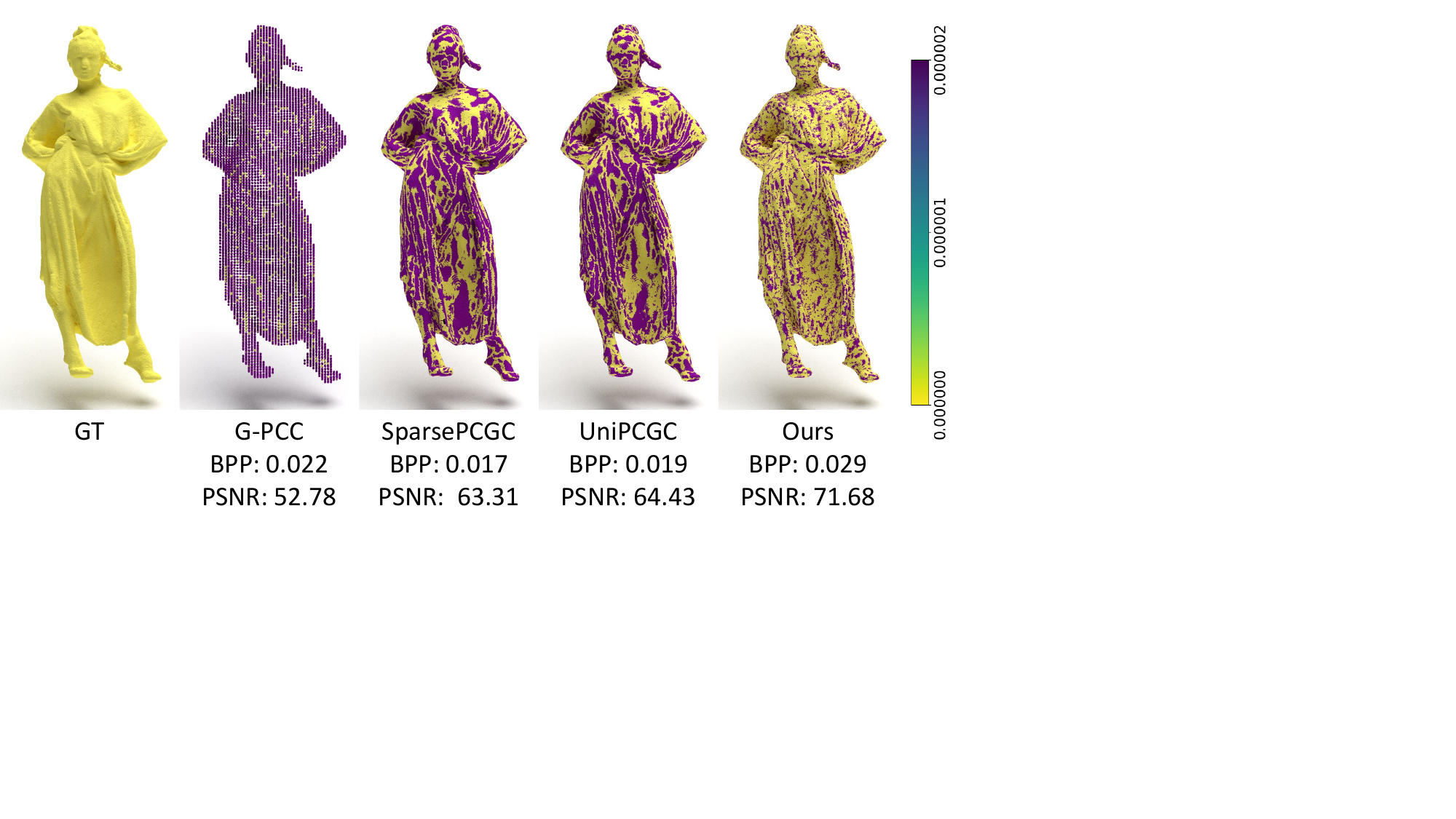}
\caption{Qualitative comparison on longdress\_vox10\_1300 from MPEG dataset. For fairness, we compare at closely matched bitrates. Errors beyond the maximum threshold are truncated. Errors are computed from the reconstructed point cloud to the ground truth point cloud and visualized on the reconstructed point cloud.}
\label{fig:mpeg_vis}
\end{figure}

\noindent\textbf{Qualitative Evaluation}. Fig.~\ref{fig:shapenet_vis} presents decompressed point clouds from the ShapeNet dataset at two bitrates, approximately 0.9 and 1.5 BPP. 
As the bitrate cannot be precisely aligned across all methods, we selected samples with the closest achievable values to ensure a fair comparison. Our approach produces a noticeably more complete reconstruction, demonstrating its ability to infer the overall structure from limited information encoded in the compressed sparse priors. 
Fig.~\ref{fig:mpeg_vis} further shows reconstruction results on \texttt{longdress\_vox10\_1300} from the MPEG dataset. 
Compared to the anchor G-PCC, our method achieves an 18.90 dB PSNR gain with only 0.007 additional bits, underscoring its strong generalization to dense point clouds.

\begin{table}[h]
\centering
\caption{Comparison of encoding and decoding time (seconds), Bits Per Point (BPP), and Chamfer Distance (CD $\times 10^{-4}$) for 10k and 32k points. $s^*$ denotes the number of DDIM denoising steps. For fairness, all methods are evaluated at closely matched Chamfer distances.}
\setlength\tabcolsep{3pt}
\normalsize
\begin{tabular}{cc|c|c|c|c c c}
 \toprule
  \multicolumn{2}{c|}{Methods} & Depoco~\cite{wiesmann2021deep} & D-PCC~\cite{he2022density}  & \multicolumn{3}{c}{Ours} \\ 
   & & & & s50 & s20 & s10 \\
\midrule
\multirow{4}{*}{10k}  &  Enc & \textbf{0.03500} & \underline{0.069} & \underline{0.069} & \underline{0.069} & \underline{0.069}\\
& Dec & \textbf{0.00072} & \underline{0.028}  & 0.32 & 0.13 & 0.07\\
& BPP & 25.24 & \underline{3.99} & \textbf{0.96} & \textbf{0.96} & \textbf{0.96}\\
& CD & \textbf{2.3} & 4.8 & \underline{3.7} & 3.8 & \underline{3.7}\\
\midrule
\multirow{4}{*}{32k} & Enc & \textbf{0.12000} & 0.35 & \underline{0.21} & \underline{0.21} & \underline{0.21}\\
& Dec & \textbf{0.00050} & \underline{0.045} & 0.74 & 0.22 & 0.14\\
& BPP & 9.08 & \underline{3.98} & \textbf{0.86} & \textbf{0.86} & \textbf{0.86}\\
& CD & 986.2 & \underline{1.9} & \textbf{1.6} & \textbf{1.6} & \textbf{1.6}\\
\bottomrule
\end{tabular}
\label{tab:runtime}
\end{table}

\begin{figure}[htbp]
\centering
\includegraphics[width=\linewidth, trim=0cm 1cm 2cm 0cm]{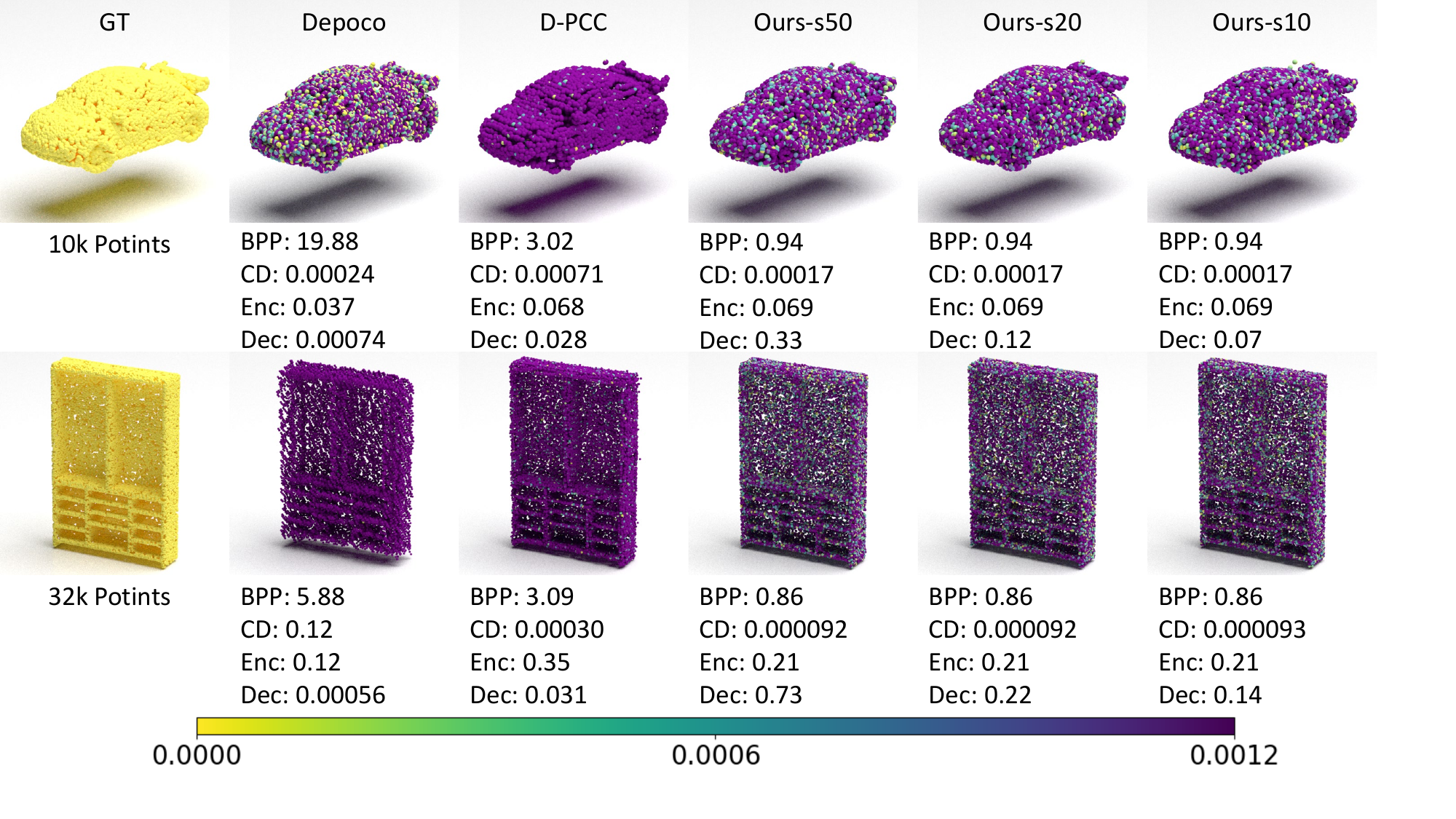}
\caption{Comparison of different denoising steps on the ShapeNet dataset with 10k and 32k points. For fairness, methods are compared at closely matched Chamfer Distance. Errors exceeding the maximum threshold are truncated. All errors are computed from the reconstructed to the ground truth point cloud and visualized on the reconstructed point cloud.}
\label{fig:ablation_vis}
\end{figure}

\noindent\textbf{Encoding/Decoding Time and Scalability Analysis.} Table~\ref{tab:runtime} and Fig.~\ref{fig:ablation_vis} analyzes the encoding/decoding time of our method. We evaluate our diffusion models with varying denoising steps against autoencoder baselines trained on 2k points and tested on 10k and 32k points. We also compare our method with the sparse-based Depoco and the general-purpose D-PCC. Under a matched Chamfer Distance, our approach consistently achieves high compression ratios both for 10k and 32k point clouds. Reducing the denoising steps from 50 to 10 yields over $4\times$ and $5\times$ faster decoding for 10k and 32k points, respectively, without compromising reconstruction quality. If an accelerated version of the diffusion model is applied, further performance improvements are expected.


\begin{table}[htbp]
\centering
\caption{Effectiveness of Sparse Priors on ShapeNet. Distribution Coded and Distribution Fused indicate whether distributions are used in the context-aware entropy model or the fusion block, respectively. 
}
\normalsize
\setlength\tabcolsep{2pt}
\begin{tabular}{c ccc| cc}
\toprule
Name & Model & \makecell{Distribution\\Predicted} & \makecell{Distribution\\Fused} & BPP & \makecell{CD\\($\times$ $10^{-3})$} \\ \midrule
M$_1$ & 1-stage & -  & -  &  2.45 & 3.7 \\
M$_2$ & 2-stage &  - & -  &  \textbf{0.89} & 3.6 \\
M$_3$ & 2-stage & \checkmark  & -& \underline{0.95} & \underline{3.2}\\
M$_4$ & 2-stage & \checkmark  & \checkmark &  0.98 & \textbf{2.8} \\
\bottomrule
\end{tabular}
\label{tab:condition}
\end{table}

\subsection{Ablation Study}\label{subsec:abl}

\noindent\textbf{Effectiveness of Sparse Priors.} Table~\ref{tab:condition} illustrates the effectiveness of the proposed sparse priors as conditions in our Probabilistic attention-based conditional denoiser (PACD). In Variant 1 (M$_1$), we remove the denoiser and directly encode the latent points, reducing their size to match that of the sparse points. Since it contains only the autoencoder, we refer to this as a 1-stage model. In Variant 2 (M$_2$), we use only sparse points as priors without predicting their distribution. In Variant 3, we predict the distribution as additional sparse priors but do not incorporate them into the entropy model. Variant 4 (M$_4$) represents our complete model. Comparing M$_2$, M$_3$, M$_4$ with M$_1$, decoupling sparse priors results in around 64\% savings in bitrates and 24\% reduction in Chamfer Distance, mainly because the 1-stage model struggles to decrease the lower bound of entropy with latent points constrained by reconstruction quality. Comparing M$_3$, M$_4$ with M$_2$, the distributions provide valuable guidance for both latent denoising and bitrate reduction during the binary coding process, with only a marginal increase ($\approx$10\%) in bits. The effectiveness of these distributions in the context-aware entropy model is further analyzed in Table~\ref{tab:abl_entropy} and in the subsequent paragraph.

\begin{table}[htbp]
\centering
\caption{Comparison of different fusion strategies. M1 and M2 are explained in the text. \checkmark in Fusion means  concatenation of the sparse points with latent points followed by MLP.
}
\setlength\tabcolsep{7pt}
\normalsize
\begin{tabular}{cccc| cc}
\toprule
\multicolumn{4}{c}{Fusion Strategy} & BPP  & CD ($\times 10^{-3}$)\\
\midrule
M$_5$        &        -          &          -        &        -         & 4.12 & \textbf{2.5} \\
M$_6$         &        -          &             -     &         -        & 1.20 & 3.6 \\ \midrule
\multicolumn{2}{c}{Downsample} & \multicolumn{2}{c|}{Upsample} & \multirow{2}{*}{BPP}  & \multirow{2}{*}{CD ($\times 10^{-3}$)} \\  
Fusion & Attn. & Fusion & Attn.  \\
\midrule
\checkmark        &        -          &     -             &        -         & 1.10 & 3.3 \\
\checkmark        & \checkmark       &         -         &        -         & 1.03 & 3.1 \\
\checkmark        & \checkmark       & \checkmark         &        -         & \underline{0.99} & 3.0 \\
\checkmark        & \checkmark       & \checkmark         & \checkmark      & \textbf{0.98} & \underline{2.8} \\
\bottomrule
\end{tabular}
\label{tab:abl_diffusion}
\end{table}

\vspace{1mm}\noindent\textbf{Effectiveness of Condition Fusion Strategy.} We evaluate different fusion strategies with a fixed $N_S=76$. As shown in Table~\ref{tab:abl_diffusion}, Variant 5 (M$_5$) replaces the multi-step denoiser with a regressive prior-guided module, while keeping the architecture unchanged and retraining the entire pipeline end-to-end. In Variant 6 (M$_6$), sparse conditions are introduced by replacing the first $N_S$ noisy latent points with sparse points within the fusion block. Note that, although the end-to-end model $M_5$ achieves nearly 10\% improvement in Chamfer Distance (CD), it requires significantly 3$\times$ extra bits compared to our latent diffusion models. The second group presents models with and without fusion and attention mechanisms applied in both the downsampling and upsampling processes. Compared the second group with M$_6$, by retaining all the noisy latents and then sampling from the combined set of latent and sparse points, the The \textit{concatenation and MLP} strategy provides a more comprehensive understanding of the overall structure, achieving at least 8\% BPP savings and an 8\% PSNR increase. Additionally, the use of probabilistic cross-attention during both downsampling and upsampling stages enhances the identification of potential points within noisy latent representations at both intra-point and inter-point levels.

\begin{table}[tbp]
\centering
\caption{Effectiveness of entropy coding strategies. 
Distribution Coded specifies whether sparse points (Pt) or features (Feat) are coded by distribution enhanced context aware entropy. Preserved Size indicates whether $Z_S$ is preserved at the same size as $Y_S$ or downsampled by a factor of 2 at the bottleneck.}
\vspace{-3pt}
\setlength\tabcolsep{3pt}
\vspace{1mm}
\normalsize
\begin{tabular}{cc cc | c c}
\toprule
\makecell{Predictor\\Baseblock} & \makecell{Distribution \\Coded} & \makecell{Preserved\\Size} & Attn. & BPP & \makecell{CD\\($\times$ $10^{-3}$)} \\
\midrule
MLP & Feat &       -     &- & 1.35 & 4.7 \\
Conv & Feat &    -        & -& 1.22 & 3.5 \\ \midrule
Conv & Pt & \checkmark &- & 1.18 & 3.4 \\ 
Conv & Feat & \checkmark & -& 1.10 & 3.4 \\
Conv & Pt & \checkmark & \checkmark & \underline{1.05} & \textbf{3.1} \\\midrule
MLP & Feat &      -      & \checkmark & 1.15 & 3.5 \\ 
Conv & Pt &       -     & \checkmark & 1.33 & 3.7 \\ 
Conv & Feat &       -     & \checkmark & 1.12 & \underline{3.2} \\ \midrule
Conv & Feat & \checkmark & \checkmark & \textbf{1.03} & \textbf{3.1} 
\\ \bottomrule
\end{tabular}
\vspace{-3mm}
\label{tab:abl_entropy}
\end{table}

\vspace{1mm}\noindent\textbf{Effectiveness of Distributions in the Entropy Coding Strategies.} Table~\ref{tab:abl_entropy} highlights the benefits of including PACD and distributions in the binary coding of sparse points in detail. In the first and third groups of the table, predicting distribution through convolutional layers outperforms through MLP by 0.13, 0.03 in BPP and 1.2, 0.3 in Chamfer Distance, which can be linked to the larger receptive field, enabling more accurate prediction of the distributions. Although point clouds are unordered, we hypothesize that the order of points influences the contextual distribution during the binary sequential point-to-point coding process. In the second group, we demonstrate that keeping the dimension of local distribution same as sparse points improves both metrics, as more detailed inter-point dependencies help identify critical key points from higher levels. Furthermore, the second and third groups demonstrate the advantages of using a context-aware entropy model for feature coding rather than point coding, as points are inherently sparse and require fewer bits, which is less than the additional storage introduced by the distributions. Conversely, feature coding balances the bit reduction with the cost of storing the distribution. The last group emphasizes the efficiency of local distribution in binary coding and PACD. 
By maintaining a consistent distribution size alongside the sparse points, binary coding efficiency is enhanced, leading to a lower bits-per-point metric.

\section{Conclusion}
We propose a decoupled sparse priors-guided compression framework, DiffCom, which addresses the inherent redundancy in latent representations by separating latent points, used for high-quality reconstruction, from sparse priors, designed for efficient storage and transmission. The sparse priors are further decoupled using a Gaussian Mixture Model (GMM) into sparse points and local distributions for accurate and explainable geometry coding. DiffCom employs a probabilistic attention-based conditional denoiser (PACD) to reconstruct latent points conditioned on the sparse priors, enabling high-quality reconstruction. Furthermore, it leverages a context-aware entropy model for efficient binary coding, minimizing data loss while preserving both geometric and semantic fidelity.

\subsection{Acknowledgments}
\noindent This research was supported by the Australian Research Council (ARC) under discovery grant project \# 240101926. Professor Ajmal Mian is the recipient of an ARC Future Fellowship Award (project \# FT210100268) funded by the Australian Government. 




\end{document}